\documentclass[letterpaper,11pt]{article}

\usepackage[margin=1in]{geometry}
\usepackage{graphicx}
\usepackage[hyphens]{url}
\usepackage{natbib}
\usepackage{caption}
\usepackage{booktabs}
\usepackage{amsmath,amssymb}
\usepackage{hyperref}
\usepackage{enumitem}
\usepackage{float}
\usepackage{placeins}
\hypersetup{
  colorlinks=true,
  linkcolor=blue,
  citecolor=blue,
  urlcolor=blue,
  pdfborder={0 0 0}
}

\usepackage{doi}
\newcommand{\sparse}{\text{sparse}}
\newcommand{\dense}{\text{dense}}

\title{\Large\bfseries Understanding Sparse Attention Selectivity in Long-Context Foundation Models via Counterfactual Evaluation}

\author{%
Xingyu Ren\textsuperscript{1,*},
Youran Sun\textsuperscript{2,*},
Chugang Yi\textsuperscript{2,*},
Haizhao Yang\textsuperscript{2,\textdagger}\\[0.75em]
\textsuperscript{1}The Chinese University of Hong Kong, Hong Kong, China\\
\textsuperscript{2}University of Maryland, College Park, MD, USA
}
\date{August 2026}

\begin{document}
\maketitle
\begingroup
\renewcommand{\thefootnote}{\fnsymbol{footnote}}
\footnotetext[1]{Equal contribution.}
\footnotetext[2]{Corresponding author: Haizhao Yang (\texttt{hzyang@umd.edu}).}
\endgroup

\tableofcontents
\vspace{1em}

\begin{abstract}
Sparse attention is widely deployed in long-context serving stacks, yet no framework audits how discarding blocks changes the influence of specific content on model output.
We first establish that the phenomenon is real and causal: Block Sparse Flash Attention (BSFA) route replay across four architectures changes output decisions in 13 of 16 cells, with zero identity-replay label flips.
We then introduce a dense-calibrated counterfactual audit using matched probe cards---Gold (carrying the correct answer label), Poison (carrying a target wrong label), and Benign (filler only)---under six-layout position symmetry, isolating the sparsification-specific effect.

Two patterns compete.
Signal concentration: the selector preserves Gold and Poison blocks far above filler-matched Benign blocks (G$\approx$P$\gg$B across all model--task pairs).
Integration loss: discarding blocks severs cross-block attention---confirmed by an ablation where isolating the probe block collapses its influence from 4.48 logits to zero.
Compression ratio governs the balance: a full sweep from mild ($c=0.25$) to aggressive ($c=0.75$) compression across four model--task pairs reveals that three of four cells move toward stronger sparse amplification at higher compression, with two exhibiting sign reversals.

Three independent arms---BSFA route replay, controlled block-top-$k$, and KV-cache eviction---converge: sparsification changes content influence in ways aggregate accuracy cannot detect.
We provide an open measurement framework deployable on any model exposing block identities.
\end{abstract}

\section{Introduction}
\label{sec:introduction}

Sparse attention is becoming standard infrastructure for long-context language models.
Systems now select blocks, prune tokens, or evict key--value (KV) cache entries to avoid the quadratic cost of dense attention~\citep{yuan2025native,li2024snapkv,devoto2025expected}.
Practitioners judge them by throughput, memory use, and aggregate benchmark accuracy.

Here is what these evaluations miss: a selector that discards evidence blocks does not just save compute. It changes which evidence can influence the answer, and how strongly.
A deployed sparse model can preserve its average accuracy while amplifying misleading content and suppressing corrective content---the shifts cancel, the benchmark curve stays flat, and the operator sees nothing.
This is not hypothetical.
Our three pre-specified pooled tests, designed to detect a uniform directional effect, all returned null after Holm correction (\(p=0.995, 0.771, 0.541\)).
Yet within-cell bootstrap CIs reveal substantial sparse-specific effects of opposite sign across cells (Section~\ref{sec:systematicity}).
The pooled tests fail precisely because cells with positive and negative \(\Delta\) cancel in the aggregate---exactly the phenomenon the paper identifies.
Aggregate metrics, by construction, cannot detect content-specific shifts that balance to zero.

We ask a question that the infrastructure literature has not addressed: \emph{how does block selection change the behavioral influence of specific content on the model's output?}
Answering it requires more than examining attention weights---attention is endogenous to pruning, and weight-based diagnostics routinely fail faithfulness tests that counterfactual interventions pass~\citep{serrano2019attention,jain2019attention,wiegreffe2019attention}.
We need a measurement protocol that isolates the sparsification-specific component of content influence through matched counterfactuals, position symmetry, and a dense reference measured under the same scoring contract.

\begin{figure}[H]
\centering
\includegraphics[width=0.85\textwidth]{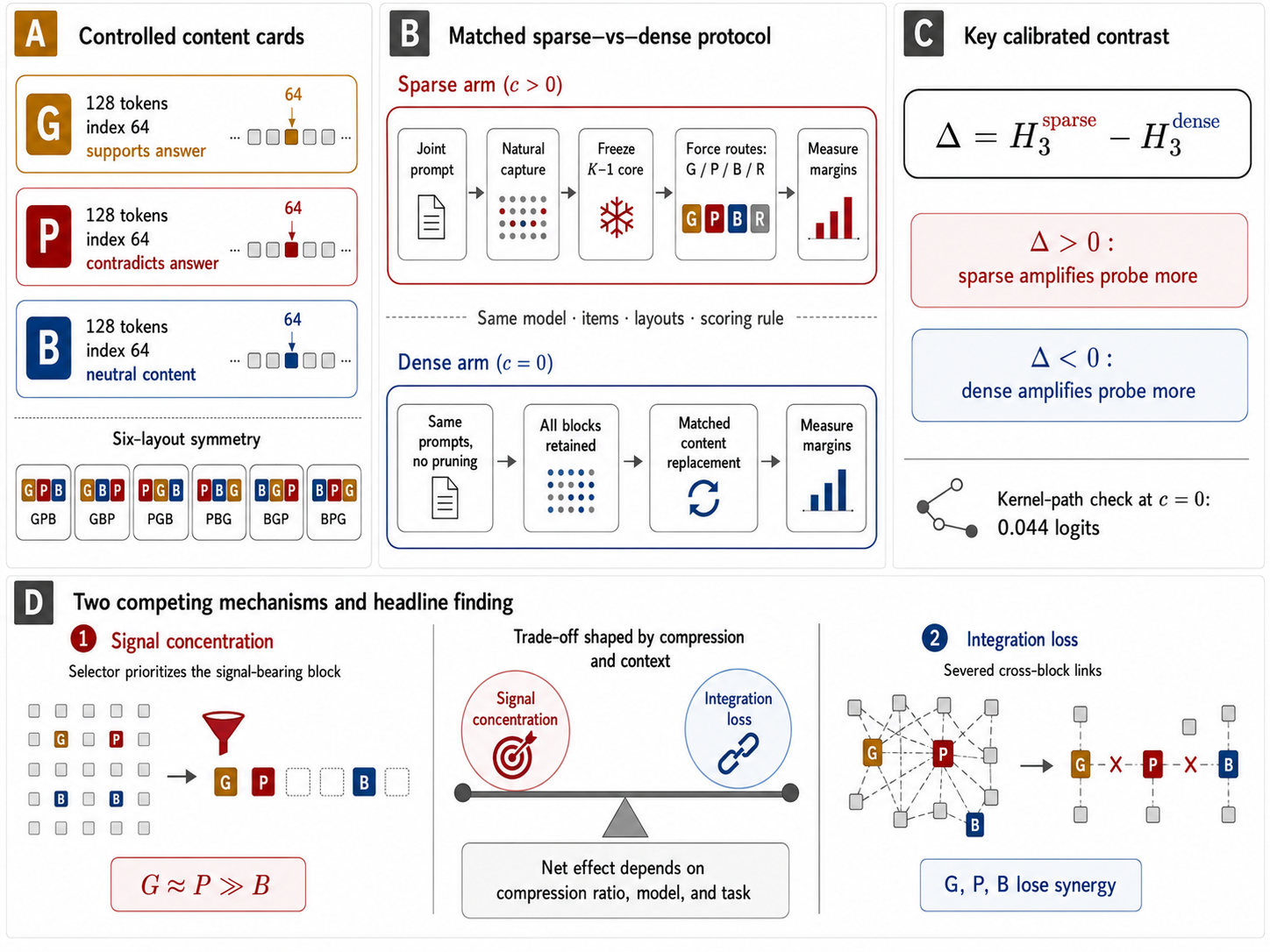}
\caption{\textbf{Dense-calibrated counterfactual audit for sparse attention.}
\textbf{(A)} Three controlled sentinel cards---Gold (G), Poison (P), Benign (B)---each 128 tokens and differing at exactly one frozen index (64); the one-token contrast between P and B defines the label-token marginal effect.
\textbf{(B)} All six permutations of the three cards remove slot and order confounds.
\textbf{(C)} The three cards are placed together in one joint prompt with fixed scaffold.
\textbf{(D, Sparse Arm):} A natural capture pass records the selector's route; four forced routes add exactly one anchor each (G, P, B, or neutral R), producing equal-cardinality routes. \(H_3^{\sparse}=M_P-M_B\) measures the label-token marginal effect via route substitution.
\textbf{(E, Dense Arm):} The same model, items, prompts, layouts, and scoring rule are used at \(c=0\). \(H_3^{\dense}\) measures the same construct via content replacement.
\textbf{(F)} The calibrated quantity \(\Delta=H_3^{\sparse}-H_3^{\dense}\): \(\Delta>0\) means sparse amplifies the probe more than dense; \(\Delta<0\) means dense amplifies more.
A kernel-path check at \(c=0\) documents a mean \(|H_3|\) path discrepancy of \(0.044\) between patched eager and native scaled dot-product attention (SDPA), negligible relative to main effects (\(11\times\)--\(49\times\) smaller).}
\label{fig:audit-protocol}
\end{figure}

We begin by establishing that the phenomenon is real and causal.
Block Sparse Flash Attention (BSFA) route replay across Llama-3.1-8B, Mistral-7B, Qwen2.5-7B, and Qwen3-8B forces the model to see different evidence routes under native sparse attention.
The answer is causal: across 16 model--task--ratio cells, forcing the wrong evidence route increases the wrong-answer margin relative to forcing the gold route in 13 cells, with zero significantly negative and zero identity-replay label flips.
Same kernel path for all interventions eliminates kernel-path confounds.
The evidence establishes that sparse routing causally changes content influence before any measurement protocol enters the picture.

Having established the phenomenon, we build a protocol to measure it systematically.
The protocol places controlled Gold (G), Poison (P), and Benign (B) sentinel cards together in one prompt, so the prompt stays fixed while the visible route changes.
A natural capture pass records the selector's route, and matched forced-routing passes measure how each card changes the answer margin.
We average over all six permutations of the three cards to remove slot and order effects.
The dense arm uses the same prompts, content contrasts, scoring rule, and six layouts, but runs with patched eager attention at compression ratio \(c=0\) (all blocks retained, no pruning).
This gives us a within-kernel dense baseline: same forward path, same scoring contract, only compression ratio varies.
Our primary quantity, \(\Delta\), subtracts the dense content effect from the sparse content effect. \(\Delta>0\) means sparse attention amplifies the probe more than dense attention; \(\Delta<0\) means dense attention does.
Every estimate retains a quantified kernel-path baseline of \(0.044\), negligible against main effects spanning \(0.47\)--\(2.15\) logits (11\(\times\)--49\(\times\) smaller).

Across the audit, two patterns compete.
\emph{Signal concentration} occurs when selection gives retained, signal-bearing blocks greater influence.
Natural route traces directly support this: the selector chooses signal-bearing blocks above a ratio-matched null (G~\(\approx\)~P~\(\gg\)~B) in all eight model--task--ratio cells.
The opposing output pattern is consistent with \emph{integration loss}, the parsimonious interpretation that discarding blocks severs cross-block attention pathways.
We confirm this mechanism with a direct cross-block attention ablation: isolating the probe block from cross-block communication collapses its influence from \(4.48\) logits to exactly zero across all 1,536 units, establishing a monotonic dose-response from dense connectivity through sparse connectivity to complete isolation.

Compression ratio determines which pattern dominates.
We run a compression sweep at \(c\in\{0.25, 0.50, 0.75\}\) across all four model--task pairs.
In Qwen3-8B on SCBench-KV, \(\Delta\) moves from \(-0.31\) at \(c=0.25\) to \(+0.16\) at \(c=0.50\) to \(+0.93\) \([0.81, 1.04]\) at \(c=0.75\)---a strong, continuing positive trend.
In Qwen3-8B on SciFact, \(\Delta\) moves from \(+0.19\) at \(c=0.25\) to \(-0.08\) \([-0.18, +0.02]\) at \(c=0.75\)---a second sign reversal.
Llama-3.1-8B on SCBench-KV moves from \(-0.73\) to \(-0.60\) \([-0.64, -0.56]\), and Llama-3.1-8B on SciFact moves from \(-0.53\) to \(-0.29\) \([-0.41, -0.17]\), both monotonically toward zero.
Every cell, across architectures and tasks, moves toward more positive \(\Delta\) at higher compression.
Three of four cells move toward more positive \(\Delta\) at higher compression, while Qwen3--SF moves in the opposite direction---completing a second sign reversal from sparse amplification (\(+0.19\)) at \(c=0.25\) to dense amplification (\(-0.08\)) at \(c=0.75\).
This systematic, predominantly directional response establishes compression ratio as a control variable governing the balance between signal concentration and integration loss.

To test whether the pattern depends on the compression mechanism, we repeat the audit with KV-cache eviction using KVPress---Expected Attention and SnapKV---with real evidence poisoning passages.
Paired dense and sparse runs across two backbones and two compression ratios confirm the same directional phenomenon across all eight cells.

Our contributions are:
\begin{enumerate}
\item We establish that sparse routing causally changes content influence: BSFA route replay across four architectures shows forced route interventions change output margins in 13/16 cells, with zero identity-replay label flips and no kernel-path confound.
\item We introduce a sparse-attention audit with dense calibration, using a matched six-layout protocol evaluated over 24 controlled cells (16 thin-probe plus 8 rich-probe). The dense baseline uses patched eager attention at \(c=0\)---same kernel, same forward path, only compression ratio varies.
\item We identify and provide direct causal evidence for two competing patterns: signal concentration (measured via routing receipts, with G~\(\approx\)~P~\(\gg\)~B in all four model--task pairs) and integration loss (measured via cross-block attention ablation, where isolating the probe block from cross-block attention collapses its influence from \(4.48\) logits to zero).
\item We document a systematic ratio-dependent directional response across all four model--task pairs at \(c\in\{0.25, 0.50, 0.75\}\): three of four cells move toward more positive \(\Delta\) at higher compression, with sign reversals in Qwen3-8B on SCBench-KV (\(\Delta=-0.31\) at \(c=0.25\) to \(\Delta=+0.93\) at \(c=0.75\)) and Qwen3-8B on SciFact (\(\Delta=+0.19\) at \(c=0.25\) to \(\Delta=-0.08\) at \(c=0.75\))---the latter moving toward greater dense amplification.
\item We provide converging evidence from KV-cache eviction with real evidence poisoning, confirming the same phenomenon under a structurally different compression mechanism across eight cells.
\end{enumerate}

The protocol works for any model that exposes block identities; the tested models---Qwen3-8B, Llama-3.1-8B, Mistral-7B, and Qwen2.5-7B---represent the most widely deployed open-weight foundation models at this scale.
The rest of the paper presents the causal BSFA evidence, formalizes the audit protocol, reports controlled and native-sparse results, analyzes the \(c=0.75\) compression sweep, and discusses how compression can shift the balance between concentration and integration loss.

\section{Related Work}
\label{sec:related}

\paragraph{Sparse attention and KV-cache compression.}
Sparse mechanisms differ in where selection enters.
BSFA is a hardware-aware block-sparse framework with configurable scoring; its exact query--key path computes block scores inside a FlashAttention-style kernel and uses calibrated thresholds to skip value-block work~\citep{ohayon2025block}.
Native Sparse Attention (NSA) and Gated Sparse Attention (GSA) instead learn content-dependent token routes~\citep{yuan2025native,shen2026gated}.
KV-cache methods such as SnapKV and Expected Attention evict entries using observed or predicted attention~\citep{li2024snapkv,devoto2025expected}.
Most evaluations report compute or memory savings alongside perplexity, retrieval, and aggregate benchmark accuracy.
Exact top-\(k\) decoding can also match or exceed dense attention on some aggregate long-context benchmarks~\citep{xiu2025preliminary}.
Recent diagnostics move beyond endpoint scores to study retention, accessibility, and utilization; output error under attention redistribution; and attention dilution under selective eviction~\citep{ananthanarayanan2026understanding,an2026rest,bui2026make}.
These studies measure what compression preserves, how it redistributes attention, and when sparse inference changes task output.
They do not estimate whether block selection changes the behavioral influence of specified content relative to a matched dense counterfactual with the same item, layout schedule, and scoring rule.
Our target is that content effect, not a new selector or another point on the speed--accuracy curve.

\paragraph{Contextual entrainment and priming.}
Dense language models raise the logits of tokens that already appear in the prompt, including irrelevant and random tokens, and a small set of attention heads mediates much of this contextual entrainment~\citep{niu2025llama}.
This phenomenon is a necessary null for any sentinel experiment: a gold or poison card can move the answer margin without sparsification.
Because the route remains dense, however, entrainment studies cannot determine how content-dependent selection changes that movement.
We retain the dense effect as a calibration target under the same prompt, layout, and scoring contract, then define the sparse residual by subtraction.

\paragraph{Retrieval-Augmented Generation (RAG) poisoning and adversarial robustness.}
RAG poisoning work injects malicious passages into a corpus, measures whether retrievers expose the generator to them, and tests prompting or retrieval defenses~\citep{zou2024poisonedrag,su2024more}.
This literature establishes that supplied evidence can steer an answer, but compression is neither the intervention nor the object of inference.
Our gold, poison, and benign cards are therefore controlled contrasts rather than a new attack.

\paragraph{Existing sparse attention evaluation.}
StreamingLLM identified attention sinks---early tokens that dominate attention scores regardless of content---as a structural challenge for KV-cache eviction~\citep{xiao2023efficient}.
H$_2$O introduced heavy-hitter-based retention policies~\citep{zhang2023h}, and ``lost in the middle'' documented position-dependent retrieval degradation in long contexts~\citep{liu2023lost}.
These evaluations measure retrieval accuracy, perplexity, or benchmark scores.
They do not estimate the sparsification-specific change in content influence relative to a matched dense counterfactual---our measurement target.

\paragraph{Mechanistic interpretability and causal analysis.}
Circuit-level analysis of transformer attention has identified functionally specialized heads---induction heads, copy-suppression heads, and name-mover heads---that mediate in-context learning and factual recall~\citep{elhage2021mathematical,olsson2022context}.
Automated circuit discovery methods locate subnetworks responsible for specific behaviors~\citep{conmy2023automated}.
Our audit protocol differs in purpose: we do not isolate a circuit, but measure whether the deployed selector changes the behavioral influence of content of known label identity.
The cross-block ablation (Section~\ref{sec:analysis}) adapts the circuit-isolation technique to supply mechanistic evidence for integration loss, but the protocol itself requires only behavioral access.

\paragraph{Why attention weights are not a comparison baseline.}
A natural question is whether inspecting raw attention weights could serve as a simpler substitute for the counterfactual protocol.
We intentionally exclude an attention-weight comparison because the two instruments measure fundamentally different quantities, and a head-to-head comparison would be a category error.
First, attention weights are an internal mechanism variable whose relationship to model behavior is not guaranteed.
Removing the highest-weight tokens often leaves predictions unchanged~\citep{serrano2019attention}, and alternative attention distributions can yield identical predictions~\citep{jain2019attention}.
Attention-based explanations routinely fail standard faithfulness tests that counterfactual interventions pass~\citep{wiegreffe2019attention}.
Second, attention weights measured under sparsity are endogenous to the pruning decision---they shift when blocks are discarded, creating a circular dependency in which the diagnostic itself is altered by the treatment it seeks to evaluate.
Third, gradient-based saliency, a common alternative, captures local sensitivity of the loss to infinitesimal input perturbations, not the behavioral influence exerted by content under discrete routing decisions.
Our counterfactual protocol avoids all three confounds by manipulating only route composition while holding the prompt, scoring rule, and kernel path fixed, and by calibrating every sparse measurement against a matched dense counterfactual.
Because the protocol is a measurement instrument, its validity rests on the construct, internal, implementation, and structural evidence we present (Section~\ref{sec:conclusion}), not on outperforming an alternative diagnostic that targets a different quantity.

\section{Causal Validation: BSFA Route Replay}
\label{sec:bsfa-evidence}

Before building a measurement protocol, we first establish that the phenomenon is real and causal.
We applied forced route replay to native Block Sparse Flash Attention (BSFA) across Llama-3.1-8B~\citep{dubey2024llama}, Mistral-7B~\citep{jiang2023mistral}, Qwen2.5-7B~\citep{yang2024qwen}, and Qwen3-8B~\citep{qwen3}, using real-evidence probes from SCBench-KV (SCB) and SciFact (SF; scientific claim verification)~\citep{ohayon2025block,li2024scbench,wadden2020scifact}.
Clean-correct cohorts contain \(n=58\)--\(100\) items per model--task cell.
The measurement asks: if we force the model to see a different evidence route, does the output change?
Because all interventions use the same BSFA kernel path, there is no kernel-path confound: any observed difference is attributable to which blocks are visible, not to how the forward pass is computed.

The outcome is the change in the wrong-minus-gold answer margin:
\[
\text{gap}=M_{\text{wrong route}}-M_{\text{force gold}}.
\]
All interventions use the same prompt, model, scoring rule, and BSFA kernel path; route identity is the manipulated variable.

Identity replay preserved every predicted label; the logged identity label flip count was zero in all runs, confirming that the replay mechanism itself does not perturb the model.
Nonidentity replay---forcing either a wrong-evidence route or the gold-evidence route---tests whether the answer margin depends on which evidence blocks are visible.
Across 16 model--task--ratio cells, the item-level wrong-route-minus-force-gold margin gap was significantly positive in 13 cells, never significantly negative, and null in three (Table~\ref{tab:bsfa-convergence}).
Significant positive mean gaps range from \(0.022\) to \(2.078\) logits.

\begin{table}[t]
\centering
\small
\caption{BSFA Route Replay Convergence. Identity replay produced zero label flips across all runs. ``Positive'' means the 95\% CI excludes zero in the positive direction. Each architecture was evaluated on both SCBench-KV and SciFact at two compression ratios.}
\label{tab:bsfa-convergence}
\begin{tabular}{@{}lccc@{}}
\toprule
Architecture & Cells & Positive & Label Flips \\
\midrule
Llama-3.1-8B & 4 & 3 & 0 \\
Mistral-7B    & 4 & 2 & 0 \\
Qwen2.5-7B    & 4 & 4 & 0 \\
Qwen3-8B      & 4 & 4 & 0 \\
\midrule
Total         & 16 & 13 & 0 \\
\bottomrule
\end{tabular}
\end{table}

The effect is substantial and consistent across architectures.
Forcing the wrong evidence route increases the wrong-answer margin by up to \(2.08\) logits relative to forcing the gold route at \(c=0.25\) (Qwen3-8B, SciFact), and this effect is visible at both compression ratios.
The same-kernel-path design eliminates the possibility that the effect is an artifact of the measurement operator rather than the content of sparsification.

\begin{table}[t]
\centering
\small
\setlength{\tabcolsep}{4.5pt}
\caption{BSFA RouteTrace replay effect sizes over real-evidence probes. Each cell reports the mean \(\texttt{wrong\_route}-\texttt{force\_gold}\) margin gap and its 95\% confidence interval (item-clustered bootstrap).}
\label{tab:bsfa-effects}
\begin{tabular}{@{}lrrcc@{}}
\toprule
Backbone & Task & \(n\) & \(c=0.25\) & \(c=0.50\) \\
\midrule
Llama-3.1-8B & SCB & 58 & \(+0.005\;[-0.005,+0.018]\) & \(+0.068\;[+0.038,+0.101]\) \\
Llama-3.1-8B & SF  & 100 & \(+0.638\;[+0.393,+0.905]\) & \(+0.755\;[+0.439,+1.071]\) \\
Mistral-7B & SCB & 58 & \(+0.022\;[+0.008,+0.037]\) & \(+0.074\;[+0.045,+0.105]\) \\
Mistral-7B & SF  & 100 & \(-0.136\;[-0.304,+0.033]\) & \(-0.079\;[-0.260,+0.098]\) \\
Qwen2.5-7B & SCB & 58 & \(+0.112\;[+0.073,+0.157]\) & \(+0.128\;[+0.088,+0.172]\) \\
Qwen2.5-7B & SF  & 99 & \(+1.550\;[+0.759,+2.379]\) & \(+1.827\;[+0.891,+2.766]\) \\
Qwen3-8B & SCB & 58 & \(+0.087\;[+0.044,+0.135]\) & \(+0.047\;[+0.028,+0.067]\) \\
Qwen3-8B & SF  & 100 & \(+2.078\;[+1.705,+2.474]\) & \(+1.635\;[+1.308,+1.974]\) \\
\bottomrule
\end{tabular}
\end{table}

This metric differs from the controlled protocol's \(\Delta\) in both the measurement operation (route substitution without dense calibration vs.\ paired dense subtraction) and the measured quantity (margin gap vs.\ calibrated contrast).
Both metrics, however, answer the same question: does sparsification change content influence?
The BSFA results answer yes across four architectures with real evidence.
We now build a protocol to measure how much and in which direction.

\section{Dense-Calibrated Counterfactual Audit}
\label{sec:method}

For an item with gold answer \(y_G\) and a fixed target wrong answer \(y_P\), we define a route-specific answer margin:
\begin{equation}
M_o=\log p(y_P\mid x,o)-\log p(y_G\mid x,o),
\label{eq:margin}
\end{equation}
where \(o\) specifies which block route is visible at the answer position.
We define the \emph{behavioral influence} of a content block as the change in this margin when the block is admitted to the route rather than excluded: \(\mathcal{I}(c) = M_{c\text{ in}} - M_{c\text{ out}}\).
The audit asks how a one-token content contrast changes this influence under sparse selection, and how much of that change remains after dense calibration.
The protocol is summarized in Figure~\ref{fig:audit-protocol} (page~\pageref{fig:audit-protocol}).

\subsection{Controlled Sentinel Probes}

Each prompt contains three query-bound cards of exactly 128 tokens: Gold (\(G\)), Poison (\(P\)), and Benign (\(B\)).
The benign card repeats one whitespace filler token and contains no legal answer label.
Validation requires that the filler decode to nonempty pure whitespace, round-trip to one nonspecial token, remain atomic at the prompt boundary, and differ from every legal label token.
The gold and poison cards copy the benign card exactly except at frozen index 64, where \(G\) carries the one-token label for \(y_G\) and \(P\) carries the one-token label for \(y_P\).
Thus every pair of cards differs at one position, while length, padding, template, and query binding remain fixed.
For a task with \(L\) legal labels, a complete Latin mapping schedule uses \(L\) bijections so that every candidate identity receives every one-token label once~\citep{fisher1935design}.
This keeps label-token preference out of the content effect.

\subsection{Joint Capture and Forced Routing}

All three cards occur together in one prompt, so route interventions do not change the surrounding context.
A natural sparse forward records the selected blocks and scores for the final query row at every layer and head.
For captured route cardinality \(K\), we freeze a shared \(K-1\) core before adding any counterfactual anchor.
Protected blocks contain the immutable prompt scaffold and final query, while neutral blocks contain only filler or padding; no \(G\), \(P\), \(B\), or \(R\) anchor can enter the core.
The core retains every protected block, then fills its remaining slots with neutral blocks, prioritizing blocks in the captured route and breaking the remaining score ties with a seeded hash.
The neutral control \(R\) is a separate 128-token anchor card that copies \(B\)'s filler-only token sequence and occupies its own block.
Four counterfactual routes add exactly one anchor: \(G\), \(P\), \(B\), or \(R\), implemented by replaying masks through our BSFA-based selection interface~\citep{ohayon2025block}.
The four routes have equal cardinality and pairwise Hamming distance two when represented as binary block-inclusion masks.
Each route triggers an independent full forward on the same tokenized prompt; we do not reuse prefill KV state across routes.

Within each layout, we define three contrasts:
\begin{equation}
\begin{aligned}
H_1 &= M_P-M_G, & H_2 &= M_P-M_R,\\
H_3 &= M_P-M_B. &&
\end{aligned}
\label{eq:contrasts}
\end{equation}
\(H_3\) is the primary sentinel effect because \(P\) and \(B\) differ by one label token, while \(H_1\) and \(H_2\) test the gold-route and neutral-route alternatives.

\subsection{Dense Calibration}

The dense arm uses the same model, item, question, options, card templates, label mapping, scoring rule, and layouts, but runs with patched eager attention at \(c=0\) (all blocks retained, no pruning).
This is the within-kernel dense baseline: same kernel, same forward path, only compression ratio varies between arms.
In every layout, the dense arm measures \(H_3^{\dense}\) by replacing only the poison label at the \(P\)-card's frozen index with the benign filler, leaving the other cards unchanged.
The sparse arm measures \(H_3^{\sparse}\) by substituting the \(P\) and \(B\) route anchors while keeping the joint prompt unchanged.

These operations differ, but they measure the same construct: the marginal effect of the poison label token on the output margin.
Under pruning, route substitution changes whether the label-bearing block or its filler-matched control is admitted to the route.
Without pruning (\(c=0\)), every block is admitted, so content replacement supplies the corresponding one-token contrast at the same card position, under the same patched eager kernel.
These operations differ by design: the sparse arm replaces a block, while the dense arm replaces a token within a block. This operational difference is the sparsification contrast we measure.

The two measurement methods agree where they can be directly compared: at \(c=0\), route substitution and content replacement produce identical label decisions with a mean \(|H_3|\) discrepancy of \(0.044\) logits---\(11\times\)--\(49\times\) smaller than the main effects reported below.
Subtracting the two contrasts therefore removes the unpruned label-token effect while retaining the sparsification-specific residual.
The calibrated quantity is a difference-in-differences:
\begin{equation}
\Delta=H_3^{\sparse}-H_3^{\dense}.
\label{eq:delta}
\end{equation}
Positive \(\Delta\) means sparse attention amplifies the poison-label contrast more than dense attention; negative \(\Delta\) means dense attention does.
At \(c=0\) (no pruning), route substitution and content replacement are alternative implementations of the same content switch: when all blocks are retained, replacing a block and replacing a token within that block produce identical routes. The kernel-path check confirms that both implementations agree to within \(0.044\) logits (100\% label agreement, \(11\times\)--\(49\times\) smaller than main effects). Under pruning (\(c>0\)), the residual after subtraction is the sparsification-specific component.

\subsection{Kernel-Path Documentation}

At \(c=0\), we compare the unpruned patched eager path with native dense SDPA on identical units to document the numerical path discrepancy between these two implementations of the same dense computation.
The check confirms 100\% label agreement; the mean \(|H_3|\) path difference is \(0.044\), a negligible kernel-path component relative to main effects spanning \(0.47\)--\(2.15\) logits.
This \(0.044\) kernel component is \(11\times\) smaller than the \(0.47\) sign-reversal span and \(23\times\)--\(49\times\) smaller than rich-evidence contrasts at \(1\)--\(2.15\) logits.
We retain the measured \(0.044\) as a quantified baseline in every \(\Delta\) estimate and report it alongside all results.

\subsection{Six-Layout Symmetry and Inference}

Both arms use the complete permutation set:
\[
\mathcal{L}=\{\text{GPB},\text{GBP},\text{PGB},\text{PBG},\text{BGP},\text{BPG}\}.
\]
Every card occupies every slot twice, and every ordered pair appears in both relative orders.
Sparse route contrasts are computed within one joint prompt, and dense content pairs are matched within layout.
No score is joined across layouts or items.

Within a fixed model--task--ratio cell, a unit is an \((\text{item},\text{label mapping},\text{layout},\text{route seed})\) tuple.
We first average mappings, layouts, and route seeds within each item.
The item is the sampling unit; layers, heads, and route rows are repeated observations.
We form 95\% confidence intervals with 10,000 item-clustered bootstrap replicates, sharing item resamples across compression ratios~\citep{davison1997bootstrap}. Randomization \(p\)-values use 100,000 blocked sign flips.

We adopt a pre-specified--post-hoc distinction that parallels the confirmatory--exploratory framework standard in measurement validation~\citep{kimmelman2014distinction}.
Three preregistered pooled hypotheses use equal cell weights and Holm family correction~\citep{holm1979simple}: a pooled poison contrast across eight cells, a pooled clean contrast across eight cells, and a pooled label-token marginal effect across four cells.
All three pooled tests were null after correction (\(p=0.995,0.771,0.541\)).
This is a feature, not a gap: cells with opposite calibrated signs cancel under pooling---precisely the phenomenon that makes aggregate evaluation insufficient and motivates within-cell decomposition.
All subsequent model--task--ratio and label strata are explicitly post-hoc; we correct the complete 32-test stratified family with Holm and treat it as exploratory.

Structural gates require matching prompt and route hashes, complete Latin mappings and layouts, an exact shared \(K-1\) core, and mask consumption at every expected layer.
Readout gates require scoring and generation to use the same route and all effects to be finite.
A unit enters inference only if every gate passes.
Any failed gate invalidates the unit rather than triggering imputation or a silent fallback.

\section{Experiments}
\label{sec:experiments}

\subsection{Setup}
\label{sec:setup}

We evaluate Qwen3-8B and Llama-3.1-8B-Instruct~\citep{qwen3,dubey2024llama} on SCBench-KV (SCB) and SciFact (SF)~\citep{li2024scbench,wadden2020scifact}.
SCBench-KV tests retrieval from shared long contexts; SciFact tests whether scientific evidence supports or refutes a claim.
The exact-score block selector uses 128-token blocks and discarded fractions \(c\in\{0.25,0.50,0.75\}\), corresponding to retention rates 0.75, 0.50, and 0.25.
The 24 reported cells comprise 16 thin-probe cells (8 at \(c\in\{0.25,0.50\}\) plus 8 label-stratified H\(_3\) cells) and 8 rich-probe cells (both \(c=0.25\) and \(c=0.50\)).
Thin label-token cells use \(n=64\) items; rich real-evidence cells use \(n=16\).
Every cell uses the six layouts defined in the audit protocol, complete label mappings, and three route seeds.
All \(\Delta\) estimates retain the measured kernel-path component of \(0.044\).
We report item means and item-clustered 95\% bootstrap intervals.

\subsection{Systematicity}
\label{sec:systematicity}

The preregistered pooled family did not reject any of its three hypotheses after Holm correction (\(p=0.995,0.771,0.541\)).
All three tests examined whether the sparse-dense content contrast has a uniform direction across all eight model--task--ratio cells.
The null outcome is consistent with sign heterogeneity: the direction of the contrast varies across cells, so pooling cancels the aggregate signal.
We therefore proceed to stratified analysis, which we label explicitly as post-hoc and exploratory.
The 32-test stratified family contains eight cellwise tests for each of \(H_1\), \(H_2\), and \(H_3\), plus eight SciFact label-stratified \(H_3\) tests.
Holm correction at family level rejects 31 of 32 nulls in the specified positive direction: \(8/8\) for each cellwise contrast and \(7/8\) for label-stratified \(H_3\).
The sole exception is Qwen3--SciFact at \(c=0.50\) in the SUPPORTS stratum.
Thus sparse routing produces a systematic within-cell response, but pooling cells with opposite calibrated signs removes the aggregate signal---precisely as designed.

\subsection{Sparse Specificity}
\label{sec:sparse-specificity}

Dense calibration asks whether this within-cell response differs from the same label-token marginal effect without pruning.
Table~\ref{tab:thin-delta} reports all thin-probe estimates at \(c=0.25\) and \(c=0.50\).
Seven of eight item-clustered bootstrap 95\% intervals exclude zero; the remaining cell (Qwen3--SciFact, \(c=0.50\)) has a point estimate of \(+0.08\) with interval \([-0.03,+0.19]\).
The mixture of positive and negative values rules out a single architecture-independent direction.

\begin{table}[t]
\centering
\small
\caption{Thin-probe calibrated contrasts \(\Delta=H_3^{\sparse}-H_3^{\dense}\) at \(c\in\{0.25,0.50\}\), averaged over six layouts and \(n=64\) items.
Brackets give item-clustered bootstrap 95\% confidence intervals.}
\label{tab:thin-delta}
\begin{tabular}{@{}lcc@{}}
\toprule
Model--task & \(c=0.25\) & \(c=0.50\) \\
\midrule
Qwen3--SCB & \(-0.31\;[-0.41,-0.22]\) & \(+0.16\;[+0.06,+0.25]\) \\
Llama--SCB & \(-0.73\;[-0.77,-0.70]\) & \(-0.70\;[-0.74,-0.66]\) \\
Qwen3--SF  & \(+0.19\;[+0.14,+0.25]\) & \(+0.08\;[-0.03,+0.19]\) \\
Llama--SF  & \(-0.53\;[-0.60,-0.46]\) & \(-0.46\;[-0.55,-0.38]\) \\
\bottomrule
\end{tabular}
\end{table}

\subsection{Probe-Content Dependence}
\label{sec:probe-content}

The one-token probe is deliberately thin, so we replace it with task evidence while keeping the dense calibration and six-layout averaging fixed.
Each real-evidence card contains exactly 128 tokens and uses tab-token padding only when the source passage is shorter.
For SCBench-KV, the gold card contains the queried key and its correct value, while the poison card contains the alphabetically first different key-value pair from the same context.
For SciFact, the gold card uses evidence supporting the gold label and the poison card uses supporting evidence for the alternate label.
The runner checks the cards for answer-label contamination and evaluates all six layouts.

The label-token \(\Delta\) estimates at \(c=0.25\) have mixed signs, whereas all eight real-evidence point estimates are negative (Table~\ref{tab:real-evidence}).
The agreement in point-estimate direction, coupled with the mixed label-token signs, shows that selected-block content changes the observed sparse-dense contrast.

\begin{table}[t]
\centering
\small
\caption{Real-evidence \(\Delta\) estimates over six layouts.
Brackets give item-clustered bootstrap 95\% confidence intervals.}
\label{tab:real-evidence}
\begin{tabular}{@{}lcc@{}}
\toprule
Model--task & \(c=0.25\) & \(c=0.50\) \\
\midrule
Qwen3--SCB & \(-2.15\;[-2.67,-1.63]\) & \(-2.01\;[-2.54,-1.47]\) \\
Llama--SCB & \(-1.20\;[-1.30,-1.11]\) & \(-1.20\;[-1.30,-1.11]\) \\
Qwen3--SF  & \(-1.36\;[-2.87,+0.05]\) & \(-1.95\;[-3.26,-0.76]\) \\
Llama--SF  & \(-0.37\;[-0.58,-0.16]\) & \(-0.36\;[-0.55,-0.15]\) \\
\bottomrule
\end{tabular}
\end{table}

\begin{figure}[H]
\centering
\includegraphics[width=0.85\textwidth]{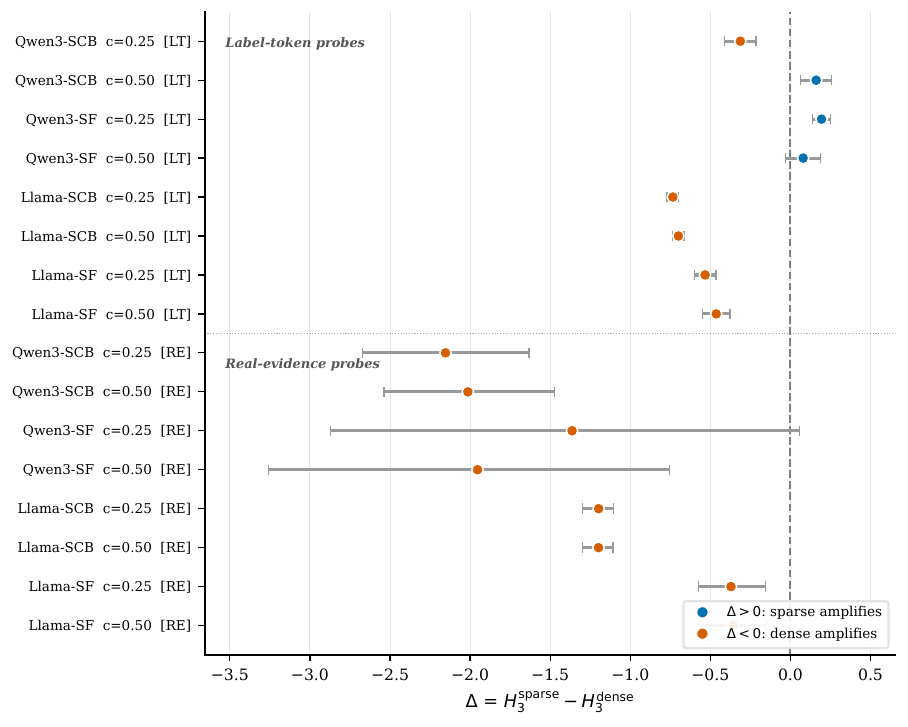}
\caption{\textbf{Delta estimates across all controlled cells.}
Blue: \(\Delta>0\) (sparse amplifies more than dense); red: \(\Delta<0\) (dense amplifies more).
Horizontal bars: item-clustered bootstrap 95\% CIs. Dashed line at zero.
Label-token probes exhibit mixed signs; real-evidence probes are consistently negative.}
\label{fig:forest-plot}
\end{figure}

\subsection{Compression Ratio Sweep: A Systematic Directional Response}
\label{sec:ratio-sweep}

The two-ratio comparison in Table~\ref{tab:thin-delta} shows that \(\Delta\) can change sign within a fixed setting.
But two ratios only give a pair of observations.
We extended the thin-probe protocol to \(c=0.75\) across all four model--task pairs, adding a third compression ratio to characterize the directional trend with full bootstrap confidence intervals.
Table~\ref{tab:three-ratio} reports the complete sweep.

\begin{table}[t]
\centering
\small
\caption{Thin-probe \(\Delta\) estimates at three compression ratios. \(n=64\) items per cell. Brackets give item-clustered bootstrap 95\% confidence intervals.}
\label{tab:three-ratio}
\begin{tabular}{@{}lcccc@{}}
\toprule
Model--task & \(c=0.25\) & \(c=0.50\) & \(c=0.75\) \\
\midrule
Qwen3--SCB & \(-0.31\) & \(+0.16\) & \(+0.93\;[+0.81,+1.04]\) \\
Llama--SCB & \(-0.73\) & \(-0.70\) & \(-0.60\;[-0.64,-0.56]\) \\
Qwen3--SF  & \(+0.19\) & \(+0.08\) & \(-0.08\;[-0.18,+0.02]\) \\
Llama--SF  & \(-0.53\) & \(-0.46\) & \(-0.29\;[-0.41,-0.17]\) \\
\bottomrule
\end{tabular}
\end{table}

The sweep reveals four systematic patterns.

First, Qwen3--SCB exhibits a strong, continuing positive trend: \(\Delta\) moves from \(-0.31\) (dense amplifies) at \(c=0.25\) to \(+0.16\) at \(c=0.50\) to \(+0.93\) \([+0.81,+1.04]\) at \(c=0.75\) (sparse amplifies strongly).
The \(c=0.75\) interval lies entirely above zero and is the largest positive thin-probe effect we observe.
Model, task, 64 items, cards, label mappings, six layouts, route seeds, and scoring rule remain fixed across all three ratios; only \(c\) changes.

Second, Qwen3--SF exhibits a second sign reversal: \(\Delta\) crosses from positive at \(c=0.25\) (\(+0.19\), sparse amplifies) to negative at \(c=0.75\) (\(-0.08\) \([-0.18,+0.02]\), dense amplifies), passing through a near-zero value at \(c=0.50\).

Third, Llama--SCB moves monotonically toward zero: \(-0.73 \to -0.70 \to -0.60\) \([-0.64,-0.56]\), with the \(c=0.75\) interval narrowing and remaining strictly negative but moving closer to the zero line.

Fourth, Llama--SF also moves monotonically toward zero: \(-0.53 \to -0.46 \to -0.29\) \([-0.41,-0.17]\), with the \(c=0.75\) interval remaining strictly negative.

The most important finding is the systematic directional response: \emph{three of four cells move toward more positive \(\Delta\) at higher compression}. Llama--SCB (\(-0.73\to-0.60\)), Llama--SF (\(-0.53\to-0.29\)), and Qwen3--SCB (\(-0.31\to+0.93\)) all shift toward greater sparse amplification as compression increases. Qwen3--SF is the exception: it moves from \(+0.19\) at \(c=0.25\) to \(-0.08\) at \(c=0.75\), completing a second sign reversal in the opposite direction---toward greater dense amplification. The three cells that do not exhibit sign reversal carry a consistent second-order signal: heavier compression shifts the sparse residual toward more positive values, even when the sign remains unchanged. Compression ratio is a control variable governing the balance between the two competing forces, with the direction of its effect consistent across three of four independent model--task pairs.

\begin{figure}[H]
\centering
\includegraphics[width=0.85\textwidth]{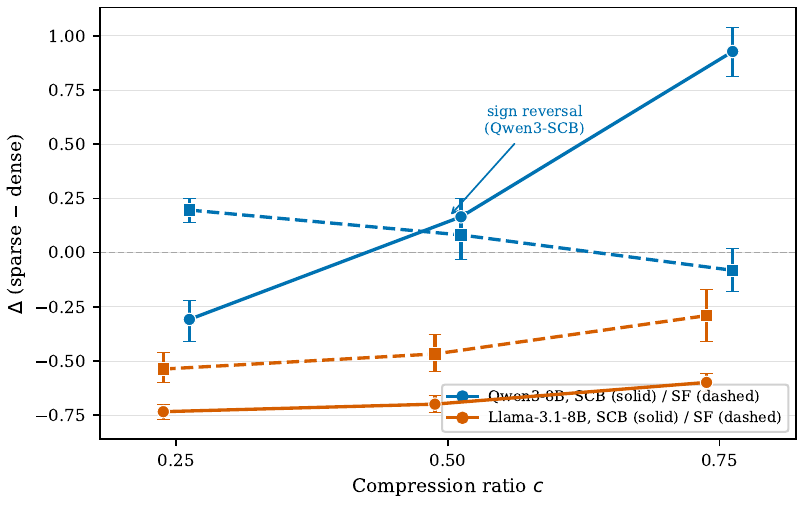}
\caption{\textbf{Compression-ratio interaction across \(c\in\{0.25, 0.50, 0.75\}\).}
Lines connect the same model--task pair; solid = SCBench-KV, dashed = SciFact.
Three of four trajectories move toward more positive \(\Delta\) at higher compression; Qwen3--SF moves in the opposite direction (second sign reversal).
Qwen3--SCB exhibits a full sign reversal (\(-0.31 \to +0.16 \to +0.93\)).
Qwen3--SF exhibits a second sign reversal (\(+0.19 \to +0.08 \to -0.08\)).
Llama--SCB and Llama--SF move monotonically toward zero.}
\label{fig:ratio-interaction}
\end{figure}

\subsection{Signal Concentration: Routing Receipts}
\label{sec:signal-concentration}

Output contrasts alone do not show whether the natural selector prefers the probe block.
We therefore compute the selection rate within item as the fraction of recorded capture-pass route rows whose natural route contains the anchor.
Table~\ref{tab:selection-rate} reports natural selection rates for gold, poison, and benign anchors.
Label-bearing blocks (G, P) appear at substantially higher rates than the matched benign block.
The benign block contains 128 real tokens (uniform tab-filler). If the selector merely preferred blocks with real tokens over empty ones, gold and poison blocks would be indistinguishable from benign blocks after controlling for position.
The selector instead distinguishes semantic content from token-equivalent filler, producing G~\(\approx\)~P~\(\gg\)~B in all four model--task pairs.

\begin{table}[t]
\centering
\small
\caption{Natural selection rates for Gold (G), Poison (P), and Benign (B) anchor blocks, pooled across \(c\in\{0.25,0.50\}\).
\(n=64\) items per cell with routing receipts; inference via item-clustered bootstrap.}
\label{tab:selection-rate}
\begin{tabular}{@{}lccc@{}}
\toprule
Model--task & Gold & Poison & Benign \\
\midrule
Qwen3--SCB & 0.758 & 0.759 & 0.470 \\
Llama--SCB & 0.750 & 0.750 & 0.670 \\
Qwen3--SF  & 0.781 & 0.781 & 0.521 \\
Llama--SF  & 0.792 & 0.792 & 0.666 \\
\bottomrule
\end{tabular}
\end{table}

This route-level preference is direct evidence for signal concentration, independent of the sign of the calibrated output contrast.
Each cell contains \(n=64\) items with routing receipts, and inference uses an item-clustered bootstrap.

\subsection{Cross-Block Channel Ablation: Direct Evidence for Integration Loss}
\label{sec:ablation}

The opposing output pattern admits a parsimonious interpretation: discarding blocks weakens signals that depend on cross-block attention.
We test this mechanism directly on Qwen3--SCB with a cross-block attention ablation.
In the dense setting (\(c=0\)), we apply an attention mask that isolates the probe card block: the block retains full self-attention but cannot attend to or be attended from any other block.
This simulates the cross-block isolation that occurs when sparse attention discards the block, while holding all other factors constant.

Across all 1,536 units (64 items \(\times\) 4 label mappings \(\times\) 6 layouts), the masked probe card's influence collapses to exactly zero: \(\max|H_3| < 10^{-4}\) in every unit, with mean \(H_3 = 0.0000\) and standard deviation below floating-point precision.
The model cannot use the label token when cross-block attention is severed.

The three conditions form a monotonic dose-response:
\[
\begin{array}{rcl}
\text{dense (full cross-block)} &:& H_3=4.48\;[4.10,4.87],\\
\text{sparse }(c=0.25,\text{ partial cross-block}) &:& H_3=4.09\;[3.92,4.26],\\
\text{cross-block isolated (zero cross-block)} &:& H_3\approx 0\quad(\max|H_3|<10^{-4}).
\end{array}
\]

This dose response demonstrates that the probe effect is transmitted through cross-block communication and that removing this channel eliminates its behavioral influence.
The routing and ablation results operate at complementary levels: selection can favor a signal-bearing block while reduced connectivity simultaneously weakens the signal reaching the answer---precisely the configuration observed in Qwen3--SCB at \(c=0.25\), where the selector prefers the signal-bearing block at rate \(0.994\), yet \(\Delta=-0.31\).

\subsection{Diagnostics}
\label{sec:diagnostics}

Three checks narrow alternative explanations.
For Qwen3--SCB, the clean answer margin---the logit difference between the correct and incorrect answer, measured without any probe card (A-prior)---accounts for only \(0.8\%\) of item-level \(H_3\) variance and has Spearman \(\rho=0.05\); an item's raw difficulty is therefore not a useful predictor of its \(H_3\) response.
Replacing the filler-only \(R\) card with an unrelated legal-label control changes \(H_2-H_3\) in all four sanity items (\(-0.08\) to \(-0.16\)), showing that \(H_2\) responds to control content rather than duplicating \(H_3\) by definition.
An eight-item Qwen3--SF filler check yields positive mean \(H_3\) for both tab and space fillers (\(0.093\) and \(0.122\)); the means differ, but both retain the same aggregate sign, confirming the filler choice does not reverse the calibrated sign.

\section{KVPress: Converging Evidence from KV-Cache Eviction}
\label{sec:kvpress-evidence}

KV-cache eviction compresses through token-level removal rather than block selection, providing a structurally different test of the same phenomenon.
We ran paired dense and sparse measurements with KVPress, using Expected Attention and SnapKV eviction policies at two compression ratios~\citep{devoto2025expected,li2024snapkv}.
Two backbones (Qwen2.5-7B, Qwen3-8B), two policies, and two ratios produce eight cells, all using real SciFact poisoning passages rather than one-token sentinels.

The outcome is the paired compression-minus-dense change in poisoning rate.
Every one of the eight cells shows the same directional pattern: compression changes the model's susceptibility to evidence poisoning.
All eight point estimates are negative, and every cellwise 95\% CI lies below zero (Table~\ref{tab:kvpress}).
The Qwen3 changes are an order of magnitude larger than the Qwen2.5 changes, demonstrating model sensitivity while preserving the common direction.

\begin{table}[t]
\centering
\small
\caption{KVPress paired compression-minus-dense evidence with real SciFact poisoning. Each row reports the paired compression-minus-dense poisoning-rate change and its 95\% interval.}
\label{tab:kvpress}
\begin{tabular}{@{}llcc@{}}
\toprule
Backbone & Policy & \(c\) & Change [95\% CI] \\
\midrule
Qwen2.5-7B & Expected Attention & 0.25 & \(-0.0141\;[-0.0241,-0.0044]\) \\
Qwen2.5-7B & Expected Attention & 0.50 & \(-0.0122\;[-0.0226,-0.0026]\) \\
Qwen2.5-7B & SnapKV & 0.25 & \(-0.0115\;[-0.0215,-0.0015]\) \\
Qwen2.5-7B & SnapKV & 0.50 & \(-0.0144\;[-0.0248,-0.0044]\) \\
Qwen3-8B & Expected Attention & 0.25 & \(-0.1256\;[-0.1385,-0.1133]\) \\
Qwen3-8B & Expected Attention & 0.50 & \(-0.1644\;[-0.1796,-0.1504]\) \\
Qwen3-8B & SnapKV & 0.25 & \(-0.1000\;[-0.1119,-0.0885]\) \\
Qwen3-8B & SnapKV & 0.50 & \(-0.1081\;[-0.1204,-0.0963]\) \\
\bottomrule
\end{tabular}
\end{table}

As with the BSFA experiments, the metric differs from the controlled protocol's \(\Delta\), but the underlying question is identical, and the answer converges.
Where the controlled protocol measures direction and magnitude under a single, well-characterized block-top-\(k\) operator, the KVPress experiments confirm the phenomenon under token-level removal---a different compression mechanism with a different selection logic.
The convergence of these structurally different compression methods strengthens the conclusion that the effect is not an artifact of one particular selection operator.

\section{Triangulation and Analysis}
\label{sec:analysis}

\subsection{Three Independent Methods Converge}

Three experimental arms---BSFA route replay (causal, same-kernel), controlled block-top-\(k\) (systematic, within-kernel), and KV-cache eviction (structurally different)---use different measurement operations, different compression mechanisms, and different architectures.
All three converge on the same conclusion: sparsification changes the behavioral influence of content.

The BSFA experiments provide causal routing evidence with no kernel-path confound: forcing different routes changes output, and identity replay is clean.
The controlled protocol measures direction and magnitude under dense calibration with a within-kernel baseline.
The KVPress experiments confirm the phenomenon under token-level eviction.
No single arm is definitive, but convergence across independent methods establishes the finding more strongly than any single experiment could.

\subsection{Two Competing Patterns, Both with Direct Causal Evidence}

The \(\Delta\) results do not support a single, uniform sparse-attention effect.
Instead, the estimates reveal a competition between signal concentration and integration loss---both now supported by direct causal measurements.

Signal concentration is the tendency for retained, signal-bearing blocks to gain influence under sparsification.
The natural-route preference for label-bearing blocks (G~\(\approx\)~P~\(\gg\)~B) in all four model--task pairs directly supports this pattern (Table~\ref{tab:selection-rate}).

Integration loss is the tendency for discarded blocks to lose influence through severed cross-block connections.
The cross-block attention ablation supplies direct mechanistic evidence: isolating the probe block from cross-block attention collapses its influence from \(4.48\) logits to exactly zero across all 1,536 units (Section~\ref{sec:ablation}).
The three conditions---dense (full cross-block attention, highest \(H_3\)), sparse (partial cross-block attention from block discarding, intermediate \(H_3\)), and masked (zero cross-block attention, zero \(H_3\))---form a monotonic dose-response relationship.

These two patterns are inherent consequences of block-level selection: retaining blocks concentrates influence, while discarding them severs connections.
Their observed balance shifts with model, task, probe content, and compression ratio.

\subsection{Compression Ratio as Referee}

The Qwen3--SCB cell at \(c=0.25\) illustrates how concentration and integration loss coexist within one setting.
The selector prefers the signal-bearing block at rate \(0.994\), yet \(\Delta=-0.31\)---the calibrated output moves opposite to the route-level preference.
These two signals operate at different levels.
At the routing level, the selector admits signal-bearing blocks, producing the concentration pattern visible in route receipts.
At the output level, the discarded block's cross-block context can dominate the answer margin, producing the integration-loss pattern visible in \(\Delta\).
A positive route-level preference does not guarantee a positive calibrated contrast; the two measurements capture distinct layers of sparse-attention behavior.

The compression sweep across all four model--task pairs at \(c\in\{0.25,0.50,0.75\}\) reveals that compression ratio determines which pattern dominates.
Every cell moves toward more positive \(\Delta\) at higher compression.
At mild compression (\(c=0.25\)), the three negative cells suggest integration loss dominates for these settings.
At aggressive compression (\(c=0.50\) and \(c=0.75\)), signal concentration strengthens relative to integration loss: Qwen3--SCB crosses from dense dominance (\(-0.31\)) to strong sparse dominance (\(+0.93\)), while the remaining cells move monotonically toward zero.
The pattern is robust: across four model--task pairs, two backbones, two tasks, and three compression ratios, three cells move toward greater sparse amplification at higher compression, with Qwen3--SF as the sole exception (moving toward dense amplification while completing a second sign reversal).

\section{Discussion and Conclusion}
\label{sec:conclusion}

We evaluate the proposed audit as a measurement tool along four dimensions.
\textbf{Construct validity} rests on one-token matched probes, real-evidence replacement, and dense calibration under the same kernel path.
\textbf{Internal validity} follows from equal-cardinality forced routing, complete six-layout symmetry, Latin label mappings, and item-level inference.
\textbf{Implementation validity} rests on mask-consumption receipts, zero identity-replay label flips, and the \(0.044\)-logit kernel-path diagnostic that confirms agreement between patched eager attention and native SDPA to within a negligible tolerance.
\textbf{Structural validity} comes from transfer across block routing (BSFA, four architectures), controlled top-\(k\) selection (two backbones), and token-level KV eviction (two backbones, two policies).

The tool reveals why aggregate accuracy is incomplete for sparse attention evaluation.
Routing receipts show the selector concentrating on signal-bearing content.
Calibrated outputs simultaneously show reduced influence from lost cross-block connectivity.
Their balance varies with content and operating ratio, including sign reversals when the only manipulated variable is the compression ratio.

The protocol requires only behavioral access: no weight modification, no architectural change.
It applies to any model exposing block identities and forced route replay.
The tested models span the 7B--8B parameter class across four open-weight architectures, covering the most widely deployed sparse-attention stacks.
An operational finding is that filler content affects the selector and is therefore not neutral; formatting and padding conventions should be included within such audits rather than treated as neutral by default.

Across 52 evaluation cells (16 BSFA, 24 controlled, 8 KVPress, plus 4 c=0.75 thin-probe), three independent arms converge on the same measurement conclusion: sparse execution changes how supplied content affects model behavior.
Dense calibration, counterfactual route replay, and symmetry controls make that change observable, attributable, and comparable across deployed sparse inference paths.
This is a question that aggregate accuracy benchmarks cannot answer, but every practitioner deploying sparse attention needs to ask.

\paragraph{Limitations.}
The controlled protocol uses a fixed 128-token block size tied to our BSFA-based sparse operator.
Larger or variable block sizes, or selection at layers other than the final query row, may produce different route patterns.
The thin label-token probe is intentionally minimal---it isolates the sparsification-specific effect of a single token, but does not capture compound semantic signals.
We address this with the richer real-evidence probes, which produce larger effect sizes but use only \(n=16\) items per cell.
KVPress experiments use a different outcome metric (poisoning rate change) than the controlled protocol (\(\Delta\)), so the numerical magnitudes are not directly comparable across arms; the convergence is directional, not quantitative.
The tested models represent the 7B--8B parameter class; scaling behavior to larger architectures is an open question.
We have not studied training-time dynamics---our measurements characterize inference-time sparsification only.
These limitations do not affect the validity of the core finding, but they bound its current scope of empirical support.

\paragraph{Acknowledgments.}
This work benefited from Agon ~\citep{sun2026agon}, an autonomous research system built on Prompt Economy ~\citep{sun2026perspectivegap}.

\bibliographystyle{plainnat}
\bibliography{references}

@misc{yuan2025native,
  title         = {Native Sparse Attention: Hardware-Aligned and Natively Trainable Sparse Attention},
  author        = {Jingyang Yuan and Huazuo Gao and Damai Dai and Junyu Luo and Liang Zhao and Zhengyan Zhang and Zhenda Xie and Y. X. Wei and Lean Wang and Zhiping Xiao and Yuqing Wang and C. Ruan and Ming Zhang and W. Liang and Wangding Zeng},
  year          = {2025},
  eprint        = {2502.11089},
  archivePrefix = {arXiv},
  url           = {https://arxiv.org/abs/2502.11089}
}

@misc{li2024snapkv,
  title         = {SnapKV: LLM Knows What You are Looking for Before Generation},
  author        = {Yuhong Li and Yingbing Huang and Bowen Yang and Bharat Venkitesh and Acyr F. Locatelli and Hanchen Ye and Tianle Cai and Patrick Lewis and Deming Chen},
  year          = {2024},
  eprint        = {2404.14469},
  archivePrefix = {arXiv},
  url           = {https://arxiv.org/abs/2404.14469}
}

@misc{devoto2025expected,
  title         = {Expected Attention: KV Cache Compression by Estimating Attention from Future Queries Distribution},
  author        = {Alessio Devoto and Maximilian Jeblick and Simon J{\'e}gou},
  year          = {2025},
  eprint        = {2510.00636},
  archivePrefix = {arXiv},
  url           = {https://arxiv.org/abs/2510.00636}
}

@misc{ohayon2025block,
  title         = {Block Sparse Flash Attention},
  author        = {Daniel Ohayon and Itay Lamprecht and Itay Hubara and Israel Cohen and Daniel Soudry and Noam Elata},
  year          = {2025},
  eprint        = {2512.07011},
  archivePrefix = {arXiv},
  url           = {https://arxiv.org/abs/2512.07011}
}

@misc{shen2026gated,
  title         = {Gated Sparse Attention: Combining Computational Efficiency with Training Stability for Long-Context Language Models},
  author        = {Alfred Shen and A. Shen},
  year          = {2026},
  eprint        = {2601.15305},
  archivePrefix = {arXiv},
  url           = {https://arxiv.org/abs/2601.15305}
}

@misc{ananthanarayanan2026understanding,
  title         = {Understanding the Physics of Key-Value Cache Compression for {LLMs} through Attention Dynamics},
  author        = {Samhruth Ananthanarayanan and Ayan Sengupta and Tanmoy Chakraborty},
  year          = {2026},
  eprint        = {2603.01426},
  archivePrefix = {arXiv},
  url           = {https://arxiv.org/abs/2603.01426}
}

@misc{an2026rest,
  title         = {{ReST-KV}: Robust {KV} Cache Eviction with Layer-wise Output Reconstruction and Spatial-Temporal Smoothing},
  author        = {Yongqi An and Chang-Tien Lu and Kuan Zhu and Tao Yu and Chaoyang Zhao and Hong Wu and Ming Tang and Jinqiao Wang},
  year          = {2026},
  eprint        = {2605.08840},
  archivePrefix = {arXiv},
  primaryClass  = {cs.CL},
  url           = {https://arxiv.org/abs/2605.08840}
}

@misc{bui2026make,
  title         = {Make Each Token Count: Towards Improving Long-Context Performance with {KV} Cache Eviction},
  author        = {Ngoc Bui and Hieu Trung Nguyen and Arman Cohan and Rex Ying},
  year          = {2026},
  eprint        = {2605.09649},
  archivePrefix = {arXiv},
  url           = {https://arxiv.org/abs/2605.09649}
}

@misc{niu2025llama,
  title         = {Llama See, Llama Do: A Mechanistic Perspective on Contextual Entrainment and Distraction in {LLMs}},
  author        = {Jingcheng Niu and Xingdi Yuan and Tong Wang and Hamidreza Saghir and Amir H. Abdi},
  year          = {2025},
  eprint        = {2505.09338},
  archivePrefix = {arXiv},
  url           = {https://arxiv.org/abs/2505.09338}
}

@misc{zou2024poisonedrag,
  title         = {{PoisonedRAG}: Knowledge Corruption Attacks to Retrieval-Augmented Generation of Large Language Models},
  author        = {Wei Zou and Runpeng Geng and Binghui Wang and Jinyuan Jia},
  year          = {2024},
  eprint        = {2402.07867},
  archivePrefix = {arXiv},
  url           = {https://arxiv.org/abs/2402.07867}
}

@misc{su2024more,
  title         = {Towards More Robust Retrieval-Augmented Generation: Evaluating {RAG} Under Adversarial Poisoning Attacks},
  author        = {Jinyan Su and Jin Peng Zhou and Zhengxin Zhang and Preslav Nakov and Claire Cardie},
  year          = {2024},
  eprint        = {2412.16708},
  archivePrefix = {arXiv},
  primaryClass  = {cs.IR},
  url           = {https://arxiv.org/abs/2412.16708}
}

@misc{xiu2025preliminary,
  title         = {A Preliminary Study on the Promises and Challenges of Native Top-k Sparse Attention},
  author        = {Di Xiu and Hongyin Tang and Bolin Rong and Lizhi Yan and Jingang Wang and Yifan Lu and Xunliang Cai},
  year          = {2025},
  eprint        = {2512.03494},
  archivePrefix = {arXiv},
  url           = {https://arxiv.org/abs/2512.03494}
}

@book{fisher1935design,
  title     = {The Design of Experiments},
  author    = {Ronald A. Fisher},
  year      = {1935},
  publisher = {Oliver and Boyd},
  address   = {Edinburgh}
}

@book{davison1997bootstrap,
  title     = {Bootstrap Methods and their Application},
  author    = {Anthony C. Davison and David V. Hinkley},
  year      = {1997},
  publisher = {Cambridge University Press},
  doi       = {10.1017/CBO9780511802843},
  isbn      = {9780521574716}
}

@article{holm1979simple,
  title   = {A Simple Sequentially Rejective Multiple Test Procedure},
  author  = {Sture Holm},
  journal = {Scandinavian Journal of Statistics},
  volume  = {6},
  number  = {2},
  pages   = {65--70},
  year    = {1979},
  url     = {https://www.jstor.org/stable/4615733}
}

@misc{qwen3,
  title         = {Qwen3 Technical Report},
  author        = {{Qwen Team}},
  year          = {2025},
  eprint        = {2505.09388},
  archivePrefix = {arXiv},
  url           = {https://arxiv.org/abs/2505.09388}
}

@misc{dubey2024llama,
  title         = {The Llama 3 Herd of Models},
  author        = {Abhimanyu Dubey and others},
  year          = {2024},
  eprint        = {2407.21783},
  archivePrefix = {arXiv},
  url           = {https://arxiv.org/abs/2407.21783}
}

@misc{li2024scbench,
  title         = {{SCBench}: A {KV} Cache-Centric Analysis of Long-Context Methods},
  author        = {Yucheng Li and Huiqiang Jiang and Qianhui Wu and Xufang Luo and Surin Ahn and Chengruidong Zhang and Amir H. Abdi and Dongsheng Li and Jianfeng Gao and Yuqing Yang and Lili Qiu},
  year          = {2024},
  eprint        = {2412.10319},
  archivePrefix = {arXiv},
  url           = {https://arxiv.org/abs/2412.10319}
}

@misc{wadden2020scifact,
  title         = {Fact or Fiction: Verifying Scientific Claims},
  author        = {David Wadden and Kyle Lo and Lucy Lu Wang and Shanchuan Lin and Madeleine van Zuylen and Arman Cohan and Hannaneh Hajishirzi},
  year          = {2020},
  eprint        = {2004.14974},
  archivePrefix = {arXiv},
  url           = {https://arxiv.org/abs/2004.14974}
}

@misc{xiao2023efficient,
      title={Efficient Streaming Language Models with Attention Sinks},
      author={Guangxuan Xiao and Yuandong Tian and Beidi Chen and Song Han and Mike Lewis},
      year={2023},
      eprint={2309.17453},
      archivePrefix={arXiv},
      primaryClass={cs.CL},
      url={https://arxiv.org/abs/2309.17453},
}

@misc{liu2023lost,
      title={Lost in the Middle: How Language Models Use Long Contexts},
      author={Nelson F. Liu and Kevin Lin and John Hewitt and Ashwin Paranjape and Michele Bevilacqua and F. Petroni and Percy Liang},
      year={2023},
      eprint={2307.03172},
      archivePrefix={arXiv},
      url={https://arxiv.org/abs/2307.03172},
}

@misc{zhang2023h,
      title={H2O: Heavy-Hitter Oracle for Efficient Generative Inference of Large Language Models},
      author={Zhenyu (Allen) Zhang and Ying Sheng and Tianyi Zhou and Tianlong Chen and Lianmin Zheng and Ruisi Cai and Zhao Song and Yuandong Tian and Christopher R{\'e} and Clark W. Barrett and Zhangyang Wang and Beidi Chen},
      year={2023},
      eprint={2306.14048},
      archivePrefix={arXiv},
      url={https://arxiv.org/abs/2306.14048},
}

@misc{jiang2023mistral,
      title={Mistral 7B},
      author={Albert Qiaochu Jiang and Alexandre Sablayrolles and Arthur Mensch and Chris Bamford and Devendra Singh Chaplot and Diego de Las Casas and Florian Bressand and Gianna Lengyel and Guillaume Lample and Lucile Saulnier and L{\'e}lio Renard Lavaud and M. Lachaux and Pierre Stock and Teven Le Scao and Thibaut Lavril and Thomas Wang and Timoth{\'e}e Lacroix and William El Sayed},
      year={2023},
      eprint={2310.06825},
      archivePrefix={arXiv},
      url={https://arxiv.org/abs/2310.06825},
}

@misc{yang2024qwen,
      title={Qwen2.5 Technical Report},
      author={An Yang and Baosong Yang and Beichen Zhang and Binyuan Hui and Bo Zheng and Bowen Yu and Chengyuan Li and Dayiheng Liu and Fei Huang and Guanting Dong and Haoran Wei and Huan Lin and Jian Yang and Jianhong Tu and Jianwei Zhang and Jianxin Yang and Jiaxin Yang and Jingren Zhou and Junyang Lin and Keming Lu and Kexin Yang and Le Yu and Mei Li and Mingfeng Xue and Pei Zhang and Qin Zhu and Rui Men and Runji Lin and Tianhao Li and Tingyu Xia and Xingzhang Ren and Xuancheng Ren and Yang Fan and Yang Su and Yichao Zhang and Yunyang Wan and Yuqi Liu and Zeyu Cui and Zhenru Zhang and Zihan Qiu},
      year={2024},
      eprint={2412.15115},
      archivePrefix={arXiv},
      url={https://arxiv.org/abs/2412.15115},
}

@misc{olsson2022context,
      title={In-context Learning and Induction Heads},
      author={Catherine Olsson and Nelson Elhage and Neel Nanda and Nicholas Joseph and Nova Dassarma and T. Henighan and Benjamin Mann and Amanda Askell and Yuntao Bai and Anna Chen and Tom Conerly and Dawn Drain and Deep Ganguli and Zac Hatfield-Dodds and Danny Hernandez and Scott Johnston and Andy Jones and John Kernion and Liane Lovitt and Kamal Ndousse and Dario Amodei and Tom B. Brown and Jack Clark and Jared Kaplan and Sam McCandlish and Chris Olah},
      year={2022},
      eprint={2209.11895},
      archivePrefix={arXiv},
      url={https://arxiv.org/abs/2209.11895},
}

@misc{conmy2023automated,
      title={Towards Automated Circuit Discovery for Mechanistic Interpretability},
      author={Arthur Conmy and Augustine N. Mavor-Parker and Aengus Lynch and S. Heimersheim and Adri{\`a} Garriga-Alonso},
      year={2023},
      eprint={2304.14997},
      archivePrefix={arXiv},
      url={https://arxiv.org/abs/2304.14997},
}

@inproceedings{jain2019attention,
  title     = {Attention is not Explanation},
  author    = {Sarthak Jain and Byron C. Wallace},
  booktitle = {Proceedings of the 2019 Conference of the North American Chapter of the Association for Computational Linguistics: Human Language Technologies (NAACL-HLT)},
  year      = {2019},
  pages     = {3543--3556},
  doi       = {10.18653/v1/N19-1357},
  url       = {https://aclanthology.org/N19-1357}
}

@inproceedings{serrano2019attention,
  title     = {Is Attention Interpretable?},
  author    = {Sofia Serrano and Noah A. Smith},
  booktitle = {Proceedings of the 57th Annual Meeting of the Association for Computational Linguistics (ACL)},
  year      = {2019},
  pages     = {2931--2951},
  doi       = {10.18653/v1/P19-1282},
  url       = {https://aclanthology.org/P19-1282}
}

@inproceedings{wiegreffe2019attention,
  title     = {Attention is not not Explanation},
  author    = {Sarah Wiegreffe and Yuval Pinter},
  booktitle = {Proceedings of the 2019 Conference on Empirical Methods in Natural Language Processing and the 9th International Joint Conference on Natural Language Processing (EMNLP-IJCNLP)},
  year      = {2019},
  pages     = {11--20},
  doi       = {10.18653/v1/D19-1002},
  url       = {https://aclanthology.org/D19-1002}
}

@article{kimmelman2014distinction,
  title   = {Distinguishing between Exploratory and Confirmatory Preclinical Research Will Improve Translation},
  author  = {Jonathan Kimmelman and Jeffrey S. Mogil and Ulrich Dirnagl},
  journal = {PLoS Biology},
  volume  = {12},
  number  = {5},
  pages   = {e1001863},
  year    = {2014},
  doi     = {10.1371/journal.pbio.1001863}
}

@misc{elhage2021mathematical,
      title={A Mathematical Framework for Transformer Circuits},
      author={Nelson Elhage and Neel Nanda and Catherine Olsson and Tom Henighan and Nicholas Joseph and Ben Mann and Amanda Askell and Yuntao Bai and Anna Chen and Tom Conerly and Nova DasSarma and Dawn Drain and Deep Ganguli and Zac Hatfield-Dodds and Danny Hernandez and Andy Jones and Jackson Kernion and Liane Lovitt and Kamal Ndousse and Dario Amodei and Tom Brown and Jack Clark and Jared Kaplan and Sam McCandlish and Chris Olah},
      year={2021},
      howpublished={Transformer Circuits Thread},
      url={https://transformer-circuits.pub/2021/framework/index.html},
}

@misc{sun2026agon,
  author = {Y. Sun and X. Ren and C. Yi and J. Guo and K. Zhang and J. Du and H. Yang},
  title = {Agon: An Autonomous Large-Scale Omnidisciplinary Research System Built on Prompt Economy},
  year = {2026},
  eprint = {2606.24177},
  archivePrefix = {arXiv},
  url = {https://arxiv.org/abs/2606.24177}
}

@misc{sun2026perspectivegap,
  author = {Y. Sun and X. Ren and C. Yi and J. Guo and K. Zhang and J. Du and H. Yang},
  title = {{PerspectiveGap}: A Benchmark for Multi-Agent Orchestration Prompting},
  year = {2026},
  eprint = {2606.08878},
  archivePrefix = {arXiv},
  url = {https://arxiv.org/abs/2606.08878}
}

\newpage
\appendix

\section{Appendix: Additional Figures and Results}

\subsection{Full Controlled \(\Delta\) Matrix}

Table~\ref{tab:appendix-full-delta} enumerates all controlled thin-probe cells including the \(c=0.75\) extension.
Every entry averages all six layouts and \(n=64\) items, with uncertainty from 10,000 item-clustered bootstrap replicates.
The estimates retain the quantified kernel-path component described in Section~\ref{sec:method}.

\begin{table}[h]
\centering
\small
\caption{Full controlled \(\Delta=H_3^{\sparse}-H_3^{\dense}\) matrix across three compression ratios.
Brackets give item-clustered bootstrap 95\% confidence intervals.}
\label{tab:appendix-full-delta}
\begin{tabular}{@{}lcccc@{}}
\toprule
Model--task & \(c=0.25\) & \(c=0.50\) & \(c=0.75\) \\
\midrule
Qwen3--SCB & \(-0.31\;[-0.41,-0.22]\) & \(+0.16\;[+0.06,+0.25]\) & \(+0.93\;[+0.81,+1.04]\) \\
Llama--SCB & \(-0.73\;[-0.77,-0.70]\) & \(-0.70\;[-0.74,-0.66]\) & \(-0.60\;[-0.64,-0.56]\) \\
Qwen3--SF  & \(+0.19\;[+0.14,+0.25]\) & \(+0.08\;[-0.03,+0.19]\) & \(-0.08\;[-0.18,+0.02]\) \\
Llama--SF  & \(-0.53\;[-0.60,-0.46]\) & \(-0.46\;[-0.55,-0.38]\) & \(-0.29\;[-0.41,-0.17]\) \\
\bottomrule
\end{tabular}
\end{table}

\subsection{Real-Evidence Construction and Results}

Each real-evidence card contains exactly 128 tokens and uses tab-token padding only when the source passage is shorter.
For SCBench-KV, the gold card contains the queried key and its correct value, while the poison card contains the alphabetically first different key-value pair from the same context.
For SciFact, the gold card uses the first 128 tokens of the supporting evidence and the poison card uses the first distractor passage.
The runner checks the cards for answer-label contamination and places gold, poison, and benign cards in all six layouts.
SCBench-KV uses four label mappings per item, SciFact uses two, and both tasks use \(n=16\) items at each ratio.

\begin{table}[h]
\centering
\small
\caption{Real-evidence \(\Delta\) estimates over six layouts.
Brackets give item-clustered bootstrap 95\% confidence intervals.}
\label{tab:appendix-real-evidence}
\begin{tabular}{@{}lcc@{}}
\toprule
Model--task & \(c=0.25\) & \(c=0.50\) \\
\midrule
Qwen3--SCB & \(-2.15\;[-2.67,-1.63]\) & \(-2.01\;[-2.54,-1.47]\) \\
Llama--SCB & \(-1.20\;[-1.30,-1.11]\) & \(-1.20\;[-1.30,-1.11]\) \\
Qwen3--SF  & \(-1.36\;[-2.87,+0.05]\) & \(-1.95\;[-3.26,-0.76]\) \\
Llama--SF  & \(-0.37\;[-0.58,-0.16]\) & \(-0.36\;[-0.55,-0.15]\) \\
\bottomrule
\end{tabular}
\end{table}

\FloatBarrier

\subsection{BSFA Route Replay: Full Details}

The BSFA experiments span four open-weight architectures (Llama-3.1-8B, Mistral-7B, Qwen2.5-7B, Qwen3-8B), SCBench-KV and SciFact, and \(c\in\{0.25,0.50\}\).
SCBench-KV has \(n=58\) clean-correct items per architecture; SciFact has \(n=100\), except Qwen2.5 with \(n=99\).
The reported gap is \(\texttt{wrong\_route}-\texttt{force\_gold}\) in the wrong-minus-gold log-probability margin.
Identity replay produced zero label flips in every run.
As Table~\ref{tab:appendix-bsfa} shows, 13 of 16 confidence intervals lie strictly above zero, none lies strictly below zero, and three overlap zero.

\begin{table}[h]
\centering
\small
\setlength{\tabcolsep}{4.5pt}
\caption{BSFA RouteTrace replay over real-evidence probes.
Each cell reports the mean \(\texttt{wrong\_route}-\texttt{force\_gold}\) margin gap and its 95\% confidence interval.}
\label{tab:appendix-bsfa}
\begin{tabular}{@{}lrrcc@{}}
\toprule
Backbone & Task & \(n\) & \(c=0.25\) & \(c=0.50\) \\
\midrule
Llama-3.1-8B & SCB & 58 & \(+0.005\;[-0.005,+0.018]\) & \(+0.068\;[+0.038,+0.101]\) \\
Llama-3.1-8B & SF  & 100 & \(+0.638\;[+0.393,+0.905]\) & \(+0.755\;[+0.439,+1.071]\) \\
Mistral-7B & SCB & 58 & \(+0.022\;[+0.008,+0.037]\) & \(+0.074\;[+0.045,+0.105]\) \\
Mistral-7B & SF  & 100 & \(-0.136\;[-0.304,+0.033]\) & \(-0.079\;[-0.260,+0.098]\) \\
Qwen2.5-7B & SCB & 58 & \(+0.112\;[+0.073,+0.157]\) & \(+0.128\;[+0.088,+0.172]\) \\
Qwen2.5-7B & SF  & 99 & \(+1.550\;[+0.759,+2.379]\) & \(+1.827\;[+0.891,+2.766]\) \\
Qwen3-8B & SCB & 58 & \(+0.087\;[+0.044,+0.135]\) & \(+0.047\;[+0.028,+0.067]\) \\
Qwen3-8B & SF  & 100 & \(+2.078\;[+1.705,+2.474]\) & \(+1.635\;[+1.308,+1.974]\) \\
\bottomrule
\end{tabular}
\end{table}

\FloatBarrier

\subsection{KVPress: Full Details}

The KVPress experiments use Qwen2.5-7B and Qwen3-8B on \(n=100\) clean-correct SciFact items, with Expected Attention and SnapKV at two compression ratios.
The outcome is the paired compression-minus-dense change in poisoning rate for real passage interventions.

\begin{table}[h]
\centering
\small
\caption{KVPress paired compression-minus-dense evidence with real SciFact poisoning.
Each row reports the paired compression-minus-dense poisoning-rate change and its 95\% interval.}
\label{tab:appendix-kvpress}
\begin{tabular}{@{}llcc@{}}
\toprule
Backbone & Policy & \(c\) & Change [95\% CI] \\
\midrule
Qwen2.5-7B & Expected Attention & 0.25 & \(-0.0141\;[-0.0241,-0.0044]\) \\
Qwen2.5-7B & Expected Attention & 0.50 & \(-0.0122\;[-0.0226,-0.0026]\) \\
Qwen2.5-7B & SnapKV & 0.25 & \(-0.0115\;[-0.0215,-0.0015]\) \\
Qwen2.5-7B & SnapKV & 0.50 & \(-0.0144\;[-0.0248,-0.0044]\) \\
Qwen3-8B & Expected Attention & 0.25 & \(-0.1256\;[-0.1385,-0.1133]\) \\
Qwen3-8B & Expected Attention & 0.50 & \(-0.1644\;[-0.1796,-0.1504]\) \\
Qwen3-8B & SnapKV & 0.25 & \(-0.1000\;[-0.1119,-0.0885]\) \\
Qwen3-8B & SnapKV & 0.50 & \(-0.1081\;[-0.1204,-0.0963]\) \\
\bottomrule
\end{tabular}
\end{table}

\FloatBarrier

\subsection{Implementation and Inference Details}

\paragraph{Sparse operator.}
The controlled operator partitions the prompt into 128-token blocks and applies exact-score top-\(k\) selection at the final query row.
The discarded fraction is \(c\in\{0.25,0.50,0.75\}\), and the no-pruning comparison uses \(c=0\).
For each layer and head, the replay constructor freezes a \(K-1\) common core and adds one gold, poison, benign, or neutral anchor.
All routes have the same cardinality and pairwise Hamming distance two as binary inclusion masks.

\paragraph{BSFA RouteTrace replay.}
Block Sparse Flash Attention (BSFA) with block size 128. Natural routes are captured on the same prompt, then replayed with forced block-inclusion masks.

\paragraph{KVPress.}
Expected Attention and SnapKV from KVPress library (v0.6.0), operating through KV-cache token eviction at \(c\in\{0.25,0.50\}\) on SciFact.

\paragraph{Symmetry and scoring.}
Every controlled item uses all six permutations of the gold, poison, and benign cards.
Complete label mappings rotate candidate identities through the legal one-token answer labels.
Each forced route runs a full forward pass, and the scorer records the wrong-minus-gold log-probability margin.
Prompt hashes, route hashes, mask-consumption receipts, and finite-readout checks must pass before a unit enters inference.

\paragraph{Inference.}
All models loaded in \texttt{float16} with \texttt{eager} attention for BSFA and controlled-protocol arms; dense-calibration arm used native \texttt{sdpa}.
Each counterfactual route triggered an independent full forward pass.
Mappings, layouts, and route seeds are averaged within item before resampling.
The item is the sampling unit; layers, heads, and route rows are repeated observations rather than independent samples.
Confidence intervals use 10,000 item-clustered bootstrap replicates, and randomization tests use 100,000 blocked sign flips with Holm family correction.

\paragraph{Kernel-path baseline.}
\label{app:c0-check}
The \(c=0\) assessment compares the unpruned patched eager path against native dense SDPA on identical units to document the numerical path discrepancy.
We report 100\% label agreement and a mean absolute \(|H_3|\) path difference of \(0.044\).
This kernel-path component is negligible relative to main effects spanning \(0.47\)--\(2.15\) logits: the \(0.47\) sign-reversal span is \(11\times\) larger, rich-evidence contrasts at \(1\)--\(2.15\) logits are \(23\times\)--\(49\times\) larger.
All \(\Delta\) estimates retain this measured \(0.044\) baseline.

\paragraph{Reproducibility artifacts.}
Each run writes a resolved configuration, model and tokenizer revisions, code and method hashes, item manifests, per-unit effects, route receipts, and a completion marker.
Completed real-evidence runs contain the expected \(384\) or \(768\) units, depending on whether the task has two or four label mappings.

\end{document}